\documentclass{article}

 \usepackage[final,nonatbib]{nips_2018}

\usepackage[utf8]{inputenc} 
\usepackage[T1]{fontenc}    
\usepackage{hyperref}       
\usepackage{url}            
\usepackage{booktabs}       
\usepackage{amsfonts}       
\usepackage{nicefrac}       
\usepackage{microtype}      
\usepackage{graphicx}
\usepackage{color,soul}	

\title{ProstateGAN: Mitigating Data Bias via Prostate Diffusion Imaging Synthesis with Generative Adversarial Networks}

\author{
  Xiaodan Hu$^{1}$, Audrey G. Chung$^{1,2}$, Paul Fieguth$^{1,2}$,\\ 
  \textbf{Farzad Khalvati$^{3,4}$, Masoom A. Haider$^{3,4}$, and Alexander Wong$^{1,2}$} \\ 
  $^{1}$Department of Systems Design Engineering, University of Waterloo, Waterloo, ON, Canada\\
  $^{2}$Waterloo Artificial Intelligence Institute, Waterloo, ON, Canada\\
$^{3}$Lunenfeld-Tanenbaum Research Institute (LTRI), Sinai Health System, ON, Canada\\
$^{4}$University of Toronto, Toronto, ON, Canada \\
}

\begin{document}

\maketitle

\begin{abstract}
Generative Adversarial Networks (GANs) have shown considerable promise for mitigating the challenge of data scarcity when building machine learning-driven analysis algorithms. Specifically, a number of studies have shown that GAN-based image synthesis for data augmentation can aid in improving classification accuracy in a number of medical image analysis tasks, such as brain and liver image analysis. However, the efficacy of leveraging GANs for tackling prostate cancer analysis has not been previously explored.  Motivated by this, in this study we introduce ProstateGAN, a GAN-based model for synthesizing realistic prostate diffusion imaging data.  More specifically, in order to generate new diffusion imaging data corresponding to a particular cancer grade (Gleason score), we propose a conditional deep convolutional GAN architecture that takes Gleason scores into consideration during the training process. Experimental results show that high-quality synthetic prostate diffusion imaging data can be generated using the proposed ProstateGAN for specified Gleason scores.
\end{abstract}

\section{Introduction}
\vspace{-0.15in}
\label{Introduction}
Deep convolutional neural networks have previously shown their power in the context of medical imaging tasks~\cite{CNN_med_image}; however, the successful training of deep networks significantly relies on large-scale and balanced datasets. Since pathological data represent only a very small portion of all available medical imaging data and many patients decline to share their data due to privacy concerns, it is difficult to obtain enough samples for certain rare medical conditions, even in rather sizeable pathological datasets. As a result, most available training datasets remain unbalanced and small, constraining the overall accuracy of learned medical imaging models. Thus, effective data augmentation strategies are of great interest to the medical imaging community.

Traditional data augmentation methods include image intensity normalization, rotation, re-scaling, cropping, flipping, and Gaussian noise injection~\cite{aug_norm,aug3,aug4}. Nevertheless, the amount of data augmented is still limited, the augmented data are often highly correlated, and the augmented data can very well be meaningless during model training if essential information is removed in cropping.  In the particular context of this paper, since prostate diffusion imaging data are relatively small in size and low-resolution, typical data augmentation methods, unfortunately, have very little to offer for improving the accuracy of deep learning-driven prostate cancer analysis.

Recently, Generative Adversarial Networks (GANs)~\cite{GAN} and their variants~\cite{DCGAN,InfoGAN} have been used to increase the performance of deep networks for medical imaging tasks, due to their ability to synthesize realistic images~\cite{med_image_synthesis,HD_med_syn,DCGAN_liver,Brain_tumor_synthesis}. 

One limitation, however, is that GANs cannot fully make use of the annotated images because they cannot control the class of the generated images. In response, many medical imaging researchers have more recently used Conditional GANs (CGANs)~\cite{CGAN}, allowing the model to be conditioned on additional information such as class labels or even data from other modalities~\cite{MR2CT,CGAN_Prostate}. However, none of the current strategies have applied GANs to synthesizing diffusion imaging data for prostate cancer analysis; rather, state-of-the-art methods have focused on image translation between magnetic resonance imaging (MRI) and computed tomography (CT), and on the analysis of liver, brain, or lung cancer. In addition, the generative network is typically trained without considering any class information, which is problematic in situations of image scarcity for certain classes. 

In this paper, we propose \textbf{ProstateGAN}, a GAN-based model for synthesizing prostate diffusion imaging data, which can be used to mitigate the data bias present in machine learning-driven prostate cancer analysis. ProstateGAN has a generator and a discriminator competing with one another:
\begin{itemize}
    \item The {\em generator} accepts random noise with embedded Gleason scores as the class information, and makes use of transposed convolutions to expand the noise samples to the synthetic prostate diffusion images. 
    \item The {\em discriminator} evaluates the generator, in which the synthetic prostate diffusion images are passed into the discriminator with Gleason scores embedded. The discriminator outputs a score between zero and one, indicating the probability of the synthetic image being real (not synthetic).
\end{itemize}
Essentially, the better the generator, the more likely that the discriminator will be fooled. At the same time, the discriminator is being trained to increase its ability to distinguish false from real content by minimizing the loss between the output score and the ground truth. The contributions of this work include the synthesis of high-quality focal prostate diffusion images using generative adversarial networks (GANs) conditioned on corresponding labels. The generated synthetic data can be used to augment and balance training sets for deep networks to improve prostate cancer classification.

\section{Methods}
\vspace{-0.15in}
\label{Methods}
\begin{figure*}[t]
\centering
    \includegraphics[width = 10cm]{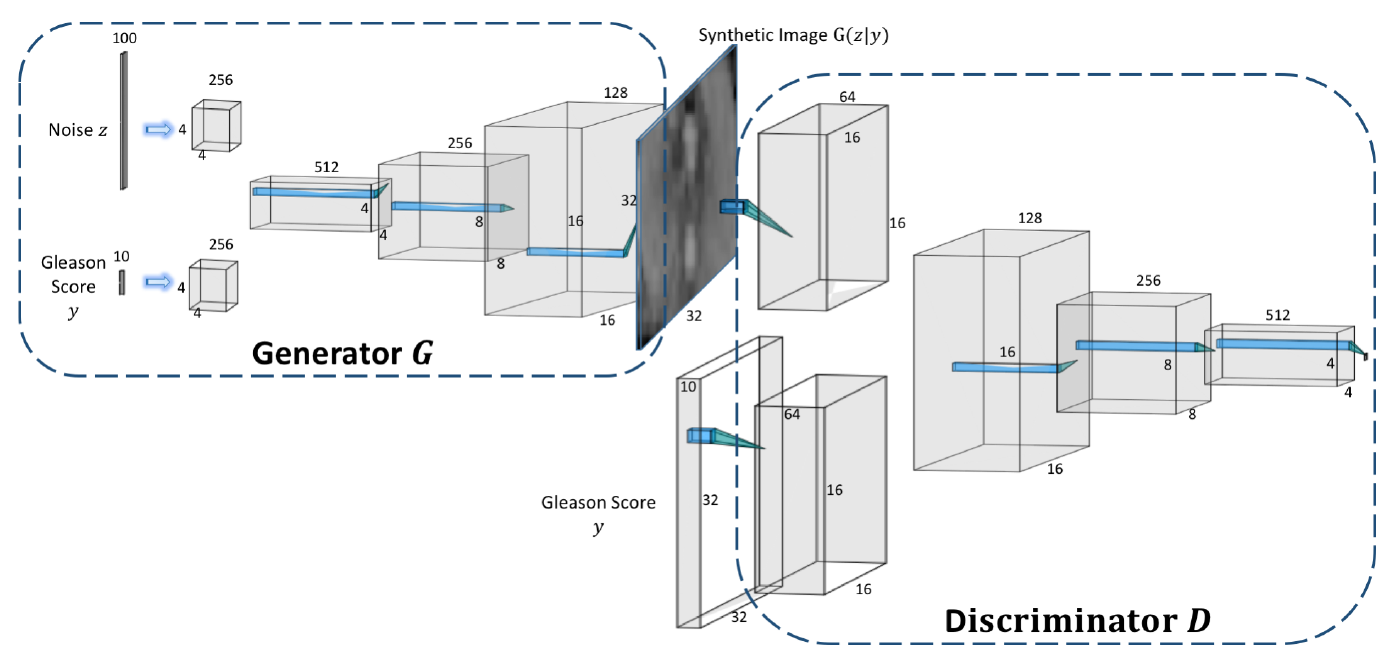}
\caption{Overview of ProstateGAN. ProstateGAN leverages a conditional deep convolutional GAN architecture that takes Gleason scores into consideration during the training process to enable the generation of synthetic prostate diffusion imaging data for specific Gleason scores.}
\label{fig_model}
\vspace{-12pt}
\end{figure*}

With the proposed ProstateGAN, we combine the ideas behind DCGAN (deep convolutional GAN)~\cite{DCGAN} and CGAN (conditional GAN)~\cite{CGAN} for the purpose of synthesizing labeled prostate diffusion imaging data, resulting in what is essentially a conditional deep convolutional GAN architecture.  In both DCGAN and CGAN, the framework consists of a generative model $G$ to estimate the data distribution $p_{x}$ from given image set $x$ and a discriminative model $D$ to attempt to differentiate synthetically generated samples from those in the training data. The objective of $G$ is to maximize the probability of $D$ being wrong, i.e., being fooled by $G$. The $G$ and $D$ of DCGAN are both deep CNNs without max pooling or fully connected layers, where $G$ uses transposed convolution for upsampling and $D$ uses strided convolutions for downsampling.  CGAN further changes $G$ and $D$ of DCGAN by adding the label $y$ as an additional parameter to both the generator to generate the corresponding images and the discriminator to distinguish real images better.

An overview of the proposed ProstateGAN is shown in Figure~\ref{fig_model}.  On the left is the generator network, taking 100 random noise samples $z$ with distribution $p_z$ (here uniform) and class label $y$, to generate a the $32 \times 32$ prostate image $\hat{x}$.  Batch normalization and ReLU are added after each of the three transposed convolutions except the last layer, which uses a {\em tanh} activation function instead. The conditional adversarial loss of $G$, i.e., what generator $G$ is trained to minimize, is defined as
\begin{equation}
\mathcal{L}_{G}^{conditional} = -\frac{1}{2} \mathbb{E}_{z \sim p_z} [log(D(G(z|y)))]
\label{lossG}
\end{equation}
where the generator $G$ is conditioned on Gleason score $y$ and $z|y$ is a joint representation.  The right part of Figure~\ref{fig_model} is the discriminator network which takes the generator output $\hat{x}$, together with the Gleason score $y$, to output the likelihood $D(\hat{x})$ indicating the probability that the input is synthetic, deciding the fidelity of $\hat{x}$. The discriminator takes the generated images $\hat{x}$ into a series of strided convolutions, with batch normalization and leaky ReLU functions applied to each layer until the last layer, replaced with a sigmoid function. The conditional loss for $D$ is defined as
\begin{equation}
\mathcal{L}_{D}^{conditional} = -\frac{1}{2} \mathbb{E}_{z \sim p_z} [1-log(D(G(z|y)))] -\frac{1}{2} \mathbb{E}_{x \sim p_x} [log(D(x|y))]
\label{lossD}
\end{equation}
where $x$ is from the true image distribution $p_x$.

\section{Results}
\vspace{-0.15in}
\label{Results}
\subsection{Experimental Setup}
\vspace{-0.1in}
Our prostate dataset consists of 1490 prostate diffusion imaging data of 104 patients, with corresponding Gleason scores (0 to 9) and PIRAD scores (2 to 5). The diffusion data were acquired using a Philips Achieva 3.0T machine at a hospital in Toronto, Ontario. Institutional research ethics board approval and written informed consent was waived by the hospital's research ethics board. For our goal of  generating realistic diffusion imaging data, which can be used for the training of improved prostate cancer classifiers, we used image-class data pairs $\{(x_i,y_i)\}_{i=1}^n$ where $x_i$ is the greyscale prostate image and $y_i \in \{0,2,...,9\}$ is the corresponding Gleason score of image $x_i$.

In our experiments, we set the epoch number to 100 and the batch size to 64. The noise samples $z$ have size 100 and are uniformly distributed. The images are first augmented by rotating, normalizing, and flipping before training. The learning rate is 0.0002 and $\beta_l$ for Adam optimizer is 0.5. Weights are initialized with the mean of 0 and standard deviation of 0.02. The slope of leaky ReLU is 0.2.

\subsection{Experimental Results}
\begin{figure*}[t]
\centering
    \includegraphics[width = 10cm]{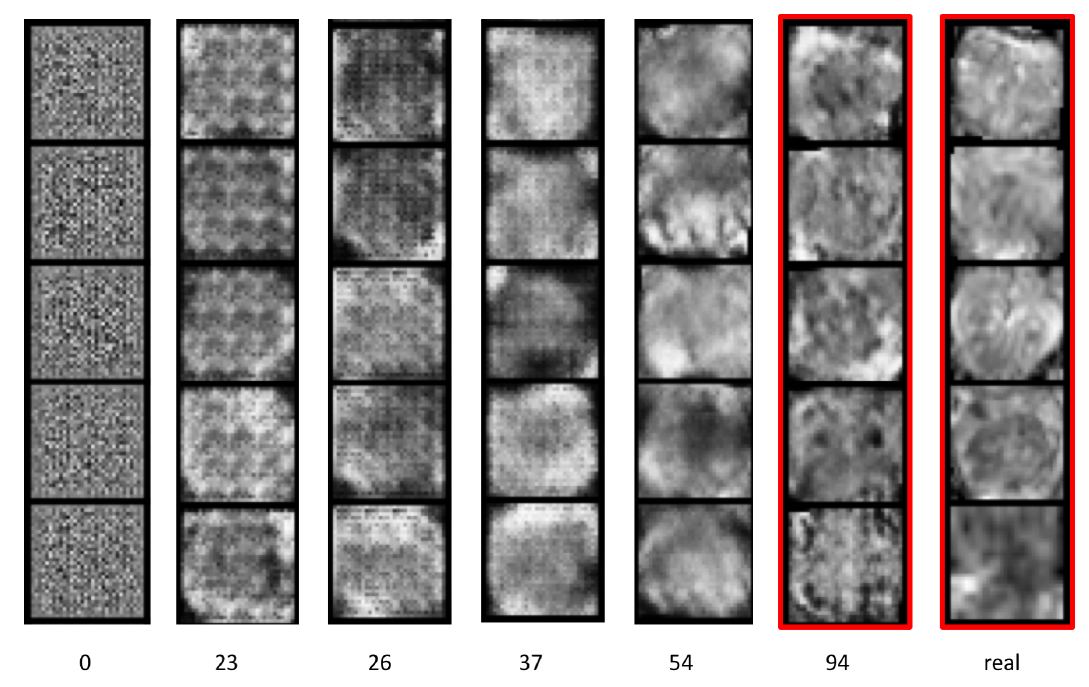}
\caption{Synthetic prostate diffusion images of the generator $G$ trained on our prostate dataset. Each column shows the synthesis output of $G$ at a given training epoch. The two columns with red boxes show the generated images and the real images of the same Gleason score, respectively.}
\label{fig_training}
\vspace{-7pt}
\end{figure*}

\vspace{-0.1in}
We trained the ProstateGAN model on our prostate dataset of diffusion imaging data and the corresponding Gleason scores. In epochs near 100, the discriminator loss converges around 1, and the generator loss converges to between 1 and 2. Figure~\ref{fig_training} shows the visualized training progress of the ProstateGAN generator, showing the synthetic images generated from $G$ for six different epochs, contrasted with real images in the right-most column, and where within each row the Gleason score is held constant. In addition, since the input noise samples $z$ for the generator are sampled before training, for each row the image of different epochs is generated using the same noise sample $\hat{z}$. It can be observed that, because $G$ uses a deconvolutional kernel of size $4 \times 4$ with a stride of 2, the generator fails to coordinate these kernels in overlapped regions, generating noise artifacts with apparent grid-lines.  In the first few epochs, these gaps inside the generated images evolve to mosaic patterns and then later fade away after dozens of epochs.

\begin{figure*}[t]
\centering
    \includegraphics[width = 10cm]{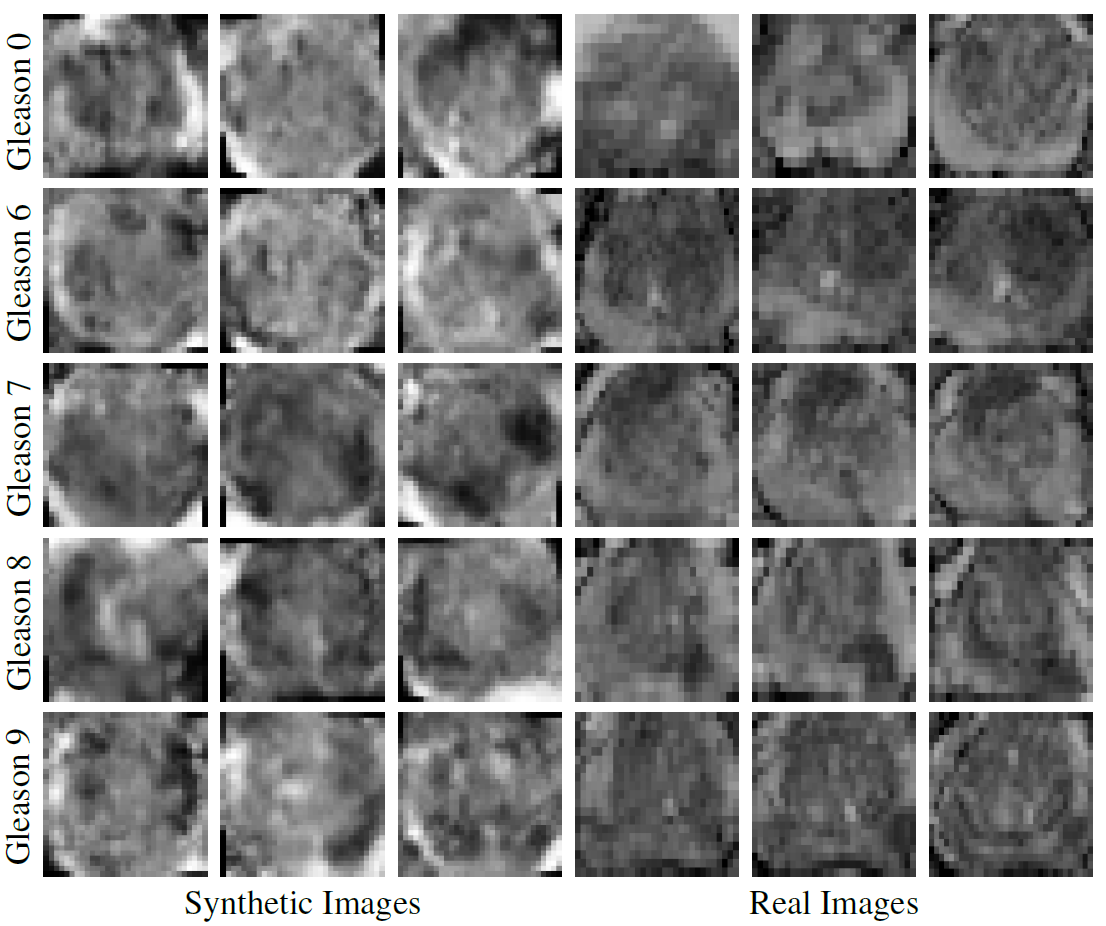}
\caption{Qualitative results (left three columns) of ProstateGAN compared with real images (right three columns).}
\label{fig_Results}
\vspace{-12pt}
\end{figure*}

The qualitative comparison between real and synthetic images is shown in Figure~\ref{fig_Results}. The generated images are, by design, of size $32 \times 32$, while the lengths and widths of the real images vary from 10 to 35. The synthetic images generated by generator $G$ have similar structure and boundaries as those of real diffusion MR images. Most notably and the goal of this work, ProstateGAN can generate synthetic images that exhibit characteristics indicative of prostate cancer, precisely the samples needed for classifier training. In particular, the abnormal darkened regions that appear in real diffusion images containing prostate cancer (i.e., Gleason scores of six or higher) can also be found in synthesized diffusion images of similar classification.

\section{Discussion}
\vspace{-0.15in}
\label{Discussion}
In this paper, we presented ProstateGAN, a model for generating realistic synthetic prostate diffusion images using a conditional deep convolutional GAN architecture, and demonstrated its ability to synthesize high-quality prostate diffusion image data by taking Gleason score into consideration during the training process. Results show that ProstateGAN can generate synthetic diffusion images corresponding to positive cancer grades (i.e., Gleason scores of six or higher) that exhibit characteristics indicative of prostate cancer. As such, ProstateGAN can potentially be used to augment and balance training datasets, an important step in mitigating data bias in prostate cancer classification.

\section*{Acknowledgement}
\vspace{-0.15in}
This work was supported by the Natural Sciences and Engineering Research Council of Canada and Canada Research Chairs Program. The authors also thank Nvidia for the GPU hardware used in this study through the Nvidia Hardware Grant Program.

\bibliographystyle{IEEEtran}
\small
\bibliography{ProstateGAN}

\end{document}